\documentclass[runningheads]{llncs}
\usepackage[T1]{fontenc}
\usepackage{graphicx}
\usepackage[table]{xcolor}
\usepackage{amsmath}  % For advanced math features
\usepackage{amssymb}  % For \mathbb
\usepackage{bm}       % For \bm

\usepackage{cite}

\usepackage{booktabs}

\begin{document}
\title{BPDec: Unveiling the Potential of Masked Language Modeling Decoder in BERT Model Pretraining}
\author{Wen Liang\inst{1}\orcidID{0009-0006-5646-2214} \and
Youzhi Liang\inst{2}}
\authorrunning{W. Liang et al.}
\titlerunning{BPDec: Masked Language Modeling Decoder.}
\institute{Google Research, Mountain View CA 94043, USA \email{liangwen@google.com} \and
Stanford University, Stanford CA 94305, USA \email{youzhil@stanford.edu}}
\maketitle              % typeset the header of the contribution
\begin{abstract}
% (Bidirectional Encoder Representations from Transformers)

BERT has revolutionized the field of natural language processing through its exceptional performance on numerous tasks. Yet, the majority of researchers have mainly concentrated on enhancements related to the model structure and pretraining, such as relative position embedding, more efficient attention mechanisms, and Masked Language Modeling, including whole word masking. In this paper, we introduce BPDec (BERT Pretraining Decoder), a novel method for modeling pretraining. BPDec is enhanced with an addition of multiple transformer blocks as a decoder, the introduction of Gradual Unmasking Attention (GUA), and the integration of output randomness for a more dynamic final output. We utilize the original BERT model as the encoder so this approach does not necessitate extensive modifications to the encoder architecture and can be seamlessly integrated into existing fine-tuning pipelines and services, offering an efficient and effective enhancement strategy with no additional training cost during the fine-tuning phase. We test multiple enhanced decoder structures after pretraining and evaluate their performance on the GLUE tasks and SQuAD tasks. Our results demonstrate that BPDec, having only undergone subtle refinements to the model structure during pretraining, significantly enhances model performance without escalating the finetuning cost, inference time and serving budget.

\keywords{Natural Language Processing  \and Language Modeling \and Foundation Model Pretraining.}

\end{abstract}

\section{Introduction}

% The journey of deep learning is marked by significant innovations with propelling advancements that redefine our understanding and capabilities in solving complex problems~\cite{bert, resnet, miamix}. ResNet revolutionized a wide range of computer vision tasks with its deep residual learning framework, enabling the training of deeper networks by addressing the vanishing gradient problem~\cite{resnet}. This breakthrough laid the groundwork for subsequent works in model architectures. The transformer model shifted the paradigm, establishing a new standard in handling sequential data~\cite{transformer}. Building on these foundations, BERT model marked another milestone in natural language processing. By pretraining on a large corpus and adopting a bidirectional approach, BERT achieved unprecedented performance across a wide range of language tasks. These contributions collectively have not only advanced the field of deep learning but also have set the stage for our current exploration into refining the BERT architecture through a pretraining and finetuning approach.

Following the groundbreaking inception of the Transformer architecture, there has been a continuous evolution in model design and methodology, further enhancing its capabilities and performance in diverse tasks. The original Transformer architecture has a separation of the encoder and decoder components, each serving a unique function in the model's processing of sequences. BERT focused exclusively on the encoder part of this architecture. By harnessing the power of the encoder, BERT excelled in creating deep, bidirectional representations with Masked Language Modeling (MLM), but it lost the sequential generation capabilities of the decoder. On the other hand, T5 (Text-to-Text Transfer Transformer) took a holistic approach by further developing both the encoder and decoder components~\cite{t5}. T5 not only embraced the full encoder-decoder structure but also innovated by treating every NLP task as a text-to-text problem and using a multi-tasking pretraining strategy. Besides, XLNet integrated the strengths of both autoregressive and autoencoding techniques~\cite{xlnet, funnel}. It further leveraged the concept of relative positional encoding, enhancing the model's ability to understand and predict the relationships between elements in a sequence.

%TransformerXL extended the Transformer's capability to process longer sequences, overcoming the limitations of the standard architecture in handling extended context~\cite{transformerxl}. This was achieved through the innovative use of relative positional encoding, allowing the model to capture dependencies over extended sequences, far beyond the absolute positional encoding that was applied in Transformer and BERT models. This enhancement not only addressed the limitations in handling extended context but also paved the way for more dynamic and contextually aware representations in sequence modeling. Complementing this, XLNet integrated the strengths of both autoregressive and autoencoding techniques~\cite{xlnet, funnel}. It further leveraged the concept of relative positional encoding, enhancing the model's ability to understand and predict the relationships between elements in a sequence. %This approach enabled XLNet to surpass BERT in numerous benchmarks. The Funnel-Transformer, another significant innovation, introduces a methodology that gradually compresses the sequence of hidden states into a shorter format, effectively reducing computation costs without compromising the model's performance across a variety of NLP tasks~\cite{funnel}. This trend underscores a growing emphasis in the field not only on achieving high accuracy but also on improving computational efficiency.

%This model incorporated varied methods such as MLM with different lengths and formats, thereby expanding the versatility and applicability of the Transformer architecture.

There has been a significant emphasis on refining strategies and methods in optimization, hyperparameter tuning, and model pretraining. For instance, RoBERTa, an iteration of BERT, has unveiled its potential by utilizing more extensive and diverse datasets, dynamic masking in Masked Language Modeling (MLM), and a refined set of hyperparameters~\cite{roberta}. This work underlines the importance of not just model design, but also the quality and variety of data and training techniques. Furthermore, some studies have delved into the masking strategy, a key factor that significantly influences the efficiency and effectiveness of the pretraining process~\cite{danqimask,spanbert}. Moreover, BERT-of-Theseus demonstrate how a compressing pretraining method can lead to more efficient models without compromising performance~\cite{theseus}.
%These strategies involve progressively replacing parts of the network with more efficient modules, thereby refining the model's structure and training process. Such developments suggest a growing recognition of the intricate interplay between various aspects of model training, extending beyond mere architectural changes to include strategic modifications in data handling, training methodologies, and structural efficiency.

DeBERTa represents a significant milestone in the ongoing evolution of language models, introducing the disentangled attention mechanism represents a significant deviation from vanilla multi-head attention~\cite{deberta}. However, compared to a BERT model with the same number of layers and similar amount of parameters, DeBERTa incurs a higher computational cost for both training and serving. Furthermore, for developers considering transitioning from BERT to DeBERTa, there is an additional development cost need to be considered, which may not always be feasible or cost-effective for every application or organization. However, DeBERTa's introduction of a pretraining-only module called the Enhanced Mask Decoder, which is essentially two additional layers discarded during fine-tuning, intrigued us more than previous architectural modifications. We were drawn to explore whether this decoder module could be generalized and optimized for use with the vanilla BERT model, and whether it could unlock encoder's ability to match or even surpass the performance of more computationally expensive models. This line of inquiry forms the basis of our research.

In this paper, we propose BPDec (BERT Pretraining Decoder), a novel architecture that harnesses the potential of the MLM (Masked Language Modeling). We introduce a novel architecture BPDec. Its encoder is identical to the vanilla BERT, but it includes a redesigned MLM decoder called BPDec. The main modifications are:
\begin{itemize}
\item Adding multiple transformer layers after the encoder to function as a decoder.
\item We proposed a new attention mechanism called Gradual Unmasking Attention (GUA) for the pretraining decoder. Conventionally, the attention mechanism for BERT is applied with a mask that prevents attending to masked positions. Instead, in Gradual Unmasking Attention, we remove this restriction and allows the model to attend to the embeddings at masked positions randomly and gradually.
\item Integration of a degree of randomness before output, where the final output is a randomly combined result of the encoder output and the decoder output.
\end{itemize}

To assess the performance of our proposed method, we conducted a series of rigorous evaluations and ablation studies across a variety of NLP tasks.
% These tasks include, but are not limited to, MNLI (Multi-Genre Natural Language Inference), SQuAD (Stanford Question Answering Dataset), RACE (Reading Comprehension from Examinations), among others. In addition, an extensive ablation study was performed to validate the effectiveness of each modification we made. This study methodically deconstructed the model to isolate and evaluate the impact of individual changes, thereby providing empirical evidence of their contribution to the model's overall performance. Furthermore, comparative analysis were conducted against baseline models, including the original BERT and other state-of-the-art architectures. These comparisons not only highlight BPDec's advancements but also offer a comprehensive understanding of its positioning in the current NLP landscape.

\section{Related Works}

\subsection{Masked Language Modeling}

In the context of traditional Language Modeling, the objective is to maximize the probability of a given sequence of tokens. This can be mathematically defined as:

\begin{equation}
\max_{\theta} \log p_{\theta}(X | \tilde{X}) = \max_{\theta} \sum_{i \in C} \log p_{\theta}(X_{i} | \tilde{X})
\end{equation}

Masked Language Modeling (MLM) has become a fundamental technique in the field of natural language processing, particularly since the introduction of the BERT model~\cite{deberta,roberta}. BERT represents a significant shift in model architecture by employing an encoder-only framework~\cite{bert}. Here, $X_{masked}$ represents the set of tokens that have been masked, and $X_{unmasked}$ is the set of tokens that remain visible to the model. The model parameters $\theta$ are optimized to maximize the log probability of correctly predicting the masked tokens given the context of unmasked tokens. BERT's MLM pretraining randomly masks 15\% of the tokens in each sequence. Within this subset, 80\% of the tokens are substituted with a [MASK] token, 10\% remain as is, and the remaining 10\% are replaced with a random token from the vocabulary. The model is trained to predict these tokens based on the context.

\begin{equation}
\max_{\theta} \log p_{\theta}(X_{masked} | \tilde{X}) = \max_{\theta} \sum_{i \in C_{\text{masked}}} \log p_{\theta}(X_{i} | X_{\text{unmasked}})
\end{equation}

\subsection{Finetuning}

Following the pretraining phase, BERT and similar models will undergo a process known as fine-tuning, which tailors them to specific tasks. In this phase, the final-layer output of the the [CLS] token, typically the first token of the sequence, is usually employed as a representative summary of the entire input sequence and is finetuned along with other model parameters to fit various downstream tasks. This fine-tuning phase is essential for adapting the generalized pretraining to more specialized applications, thereby enhancing the model's performance on task-specific datasets.

Some approaches suggest the possibility of dropping certain layers during fine-tuning for efficiency~\cite{droplayers}. Notable findings in this area highlight that:
\begin{enumerate}
\item Up to 50\% of pretrained transformer blocks can be pruned while retaining approximately 98\% efficiency in specific downstream tasks.
\item The retention of lower layers is essential for maintaining performance in downstream applications.
\item For certain tasks, a minimal subset of layers, as few as 3 out of 12, can sustain performance within a 1\% variance of the full model's capacity.
\end{enumerate}

DeBERTa introduces an approach, where it is pretrained with an additional decoder and then this decoder is dropped during the fine-tuning phase~\cite{deberta}. Similarly, the T5 model can be adapted to similar finetuning tasks by utilizing only its encoder part~\cite{t5}. Besides, BERT-of-Theseus presents a novel compression approach for BERT by progressively replacing original modules with compact substitutes during pretraining~\cite{theseus}. This method enhances interaction between original and compact models without introducing extra loss functions, ultimately outperforming existing knowledge distillation techniques on the GLUE benchmark.

\section{Methods}

Considering the structural modifications of DeBERTa and their impact on the overall computational cost of the model, we find the concept of Enhanced Mask Decoders (EMD) to be particularly intriguing. The MLM decoder offers a module that is utilized exclusively during the pretraining phase, thereby avoiding additional computational burdens during the finetuning and serving. The primary intent of introducing the Enhanced Mask Decoder (EMD) in the DeBERTa model, was to integrate absolute position information before the output layer. However, insights from other studies suggest an alternative perspective~\cite{droplayers}. The ability to understand the input text and the ability to distill language into specific embeddings are already well-established in the earlier layers, which indicates that using later layer during BERT finetuning might be unnecessary.

In related works, we observe that models like T5 and BERT-of-Theseus demonstrate that variations between pretraining and finetuning architectures do not degrade the model's capabilities, particularly in the encoder part. On the contrary, these variations can potentially enhance performance~\cite{t5, theseus}. Building upon these insights, we hypothesize that carefully designed differences in these architectures can not only avoid performance degradation but may even enhance some part of the model. This leads us to believe that, with further research and meticulous design, the original BERT model itself has untapped potential for significant improvement. Motivated by these insights and our analysis, we propose the BERT Pretraining MLM Decoder, termed BPDec, a novel architecture that serves as an innovative paradigm for BERT pretraining.

\subsection{MLM Decoder}

Our approach with BPDec maintains the fundamental encoder architecture of the original BERT, but supplements it with a specially designed decoder, which is used exclusively during pretraining. This additional decoder, consisting of transformer blocks similar to those in the encoder, is crafted to enhance the BERT encoder's ability in understanding languages. By incorporating this pretraining-only decoder, we aim to improve the overall performance of the BERT encoder without imposing significant computational costs during the subsequent phases of finetuning and deployment. The BPDec is strategically positioned after the encoder and just before the output layers for masked token prediction. It specifically sharpens the model’s ability to predict masked tokens, enriching the encoded representations with greater depth.

\subsection{Gradual Unmasking Attention (GUA)}

\begin{figure}[ht]
  \includegraphics[width=120mm]{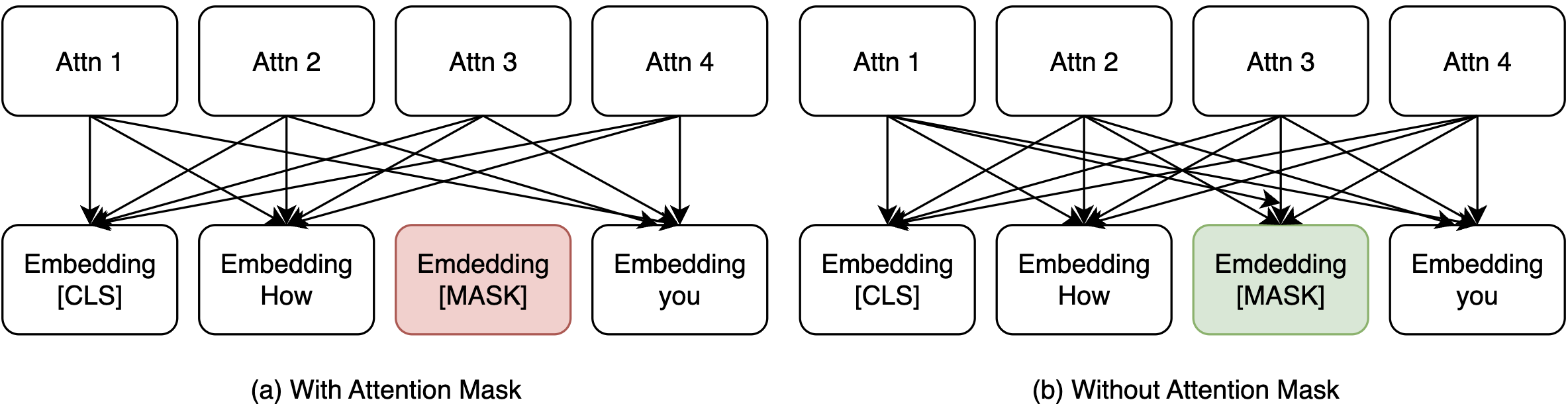}
  \centering
  \caption{Examples of attention heads with and without attention masks. (a) Attention heads will not attend to the masked embedding highlighted in red due to the attention mask. (b) The attention mask is disabled.}
  \label{fig:attn}
\end{figure}

\begin{figure}[ht]
  \includegraphics[width=120mm]{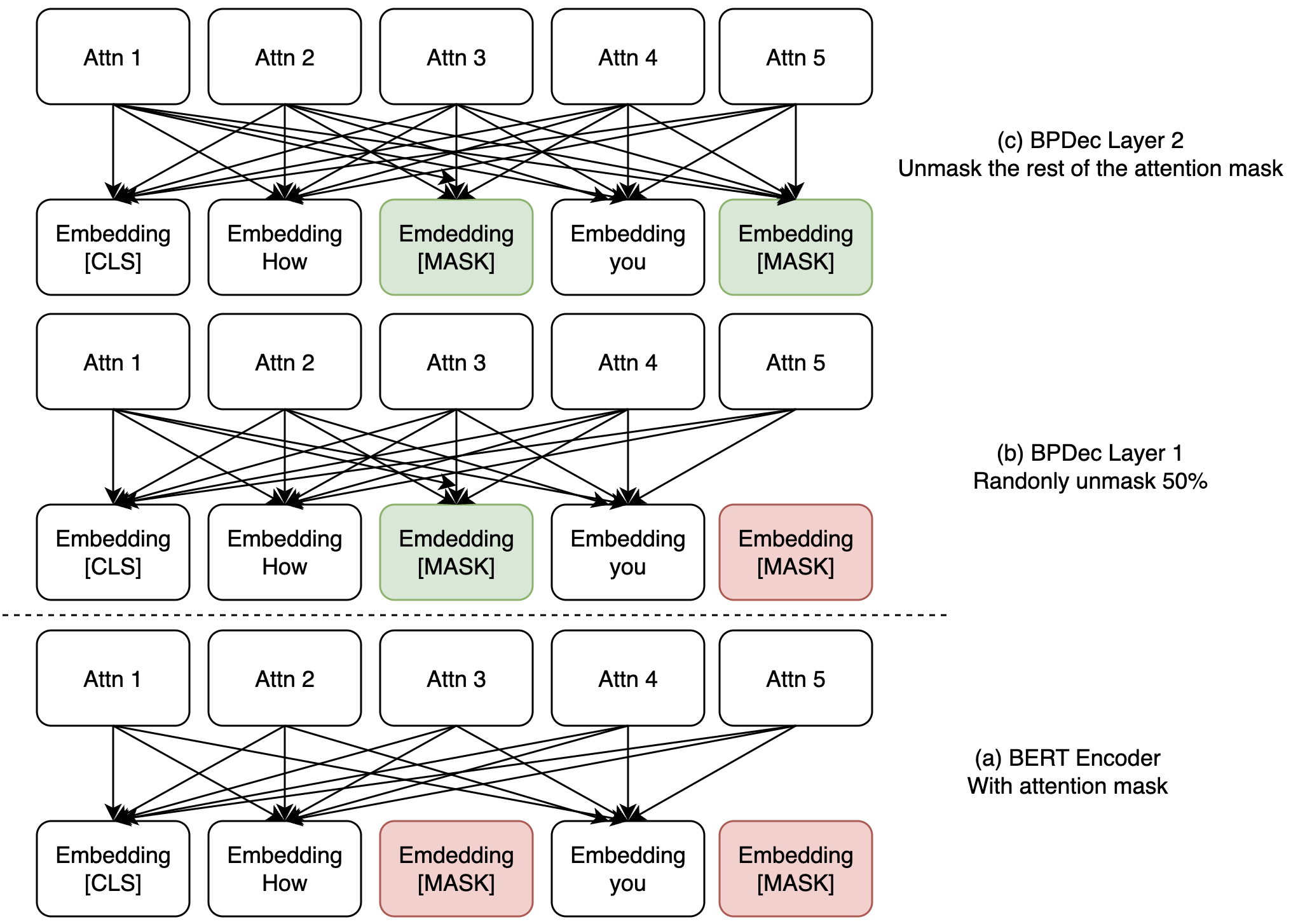}
  \centering
  \caption{Examples of gradual relaxation of attention mask. (a) Encoder layers are unaffected. (b) Randomly unmask a portion of the masked positions. (c) In the final layer of the BPDec, we unmask the rest of the attention mask and fully disable the attention mask.}
  \label{fig:attn_layers}
\end{figure}

Another key aspect of our proposed methodology in BPDec involves modifying the restrictions present in the standard multi-head attention mechanism of BERT, particularly in relation to masked positions. As shown in Figure~\ref{fig:attn}, the BERT architecture avoids attending to embeddings at the 15\% masked positions during the multi-head attention by setting an attention mask to them during pretraining. The embeddings at masked positions are considered less informative and less accurate than those are not masked. The mechanism helps ensure that the model is not unduly influenced by those masked and noised inputs.

In the development of our new pretraining strategy, we introduce the Gradual Unmasking Attention (GUA) within our BPDec. This pioneering modification challenges the conventional constraints of the attention mechanism used in BERT model pretraining. Our approach is based on the insight that as the model's layers progress deeper, the resulting output features become more abstract. We hypothesize that these abstracted features at masked positions are sufficiently accurate and detailed to serve as effective hypothesized contexts. GUA progressively unveils the attention mechanism to those hypothesized contexts of the masked inputs, fostering a more nuanced and robust formulation of language understanding. This method begins with partially unmasking the attention mask in the initial layers of the decoder. For instance, in a decoder with two transformer blocks as shown in Figure~\ref{fig:attn_layers},  the Gradual Unmasking Attention (GUA) in the first decoder layer would randomly unmask 50\% of the masked positions. This introduces a controlled amount of noise and hypothesized contexts. In the second GUA, the remaining 50\% of the positions are unveiled, allowing the model to integrate the full context. This phased approach of exposure ensures that the model gradually adapts to the changes in the inputs without being overwhelmed by sudden floods of unmasked data. More importantly, allowing the model to progressively form a complete picture of the input data significantly improves its interpretative capabilities, enhancing its ability to generate accurate and contextually relevant predictions.

\subsection{Random Mix of Encoder and Decoder Outputs}

\begin{figure}[ht]
  \includegraphics[width=60mm]{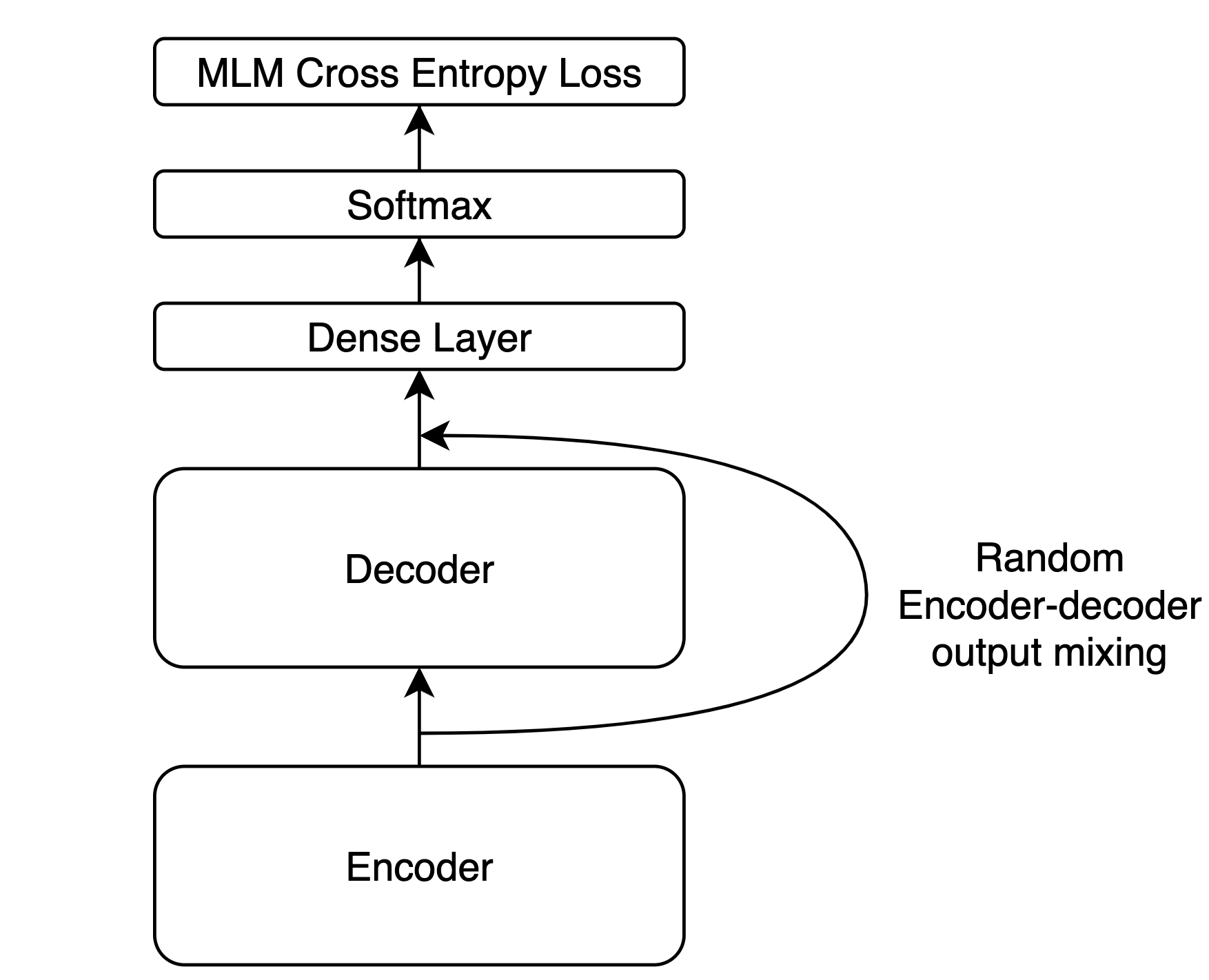}
  \centering
  \caption{Random encoder-decoder output mixing.}
  \label{fig:mixing}
\end{figure}

In our approach, we introduce a modest blending of the encoder's output into the decoder's output as shown in Figure~\ref{fig:mixing}. This technique serves a dual purpose. Firstly, it encourages the MLM task to not depend exclusively on the representations formed in the deeper layers of the decoder. Instead, it also takes into account the more direct and less processed information from the encoder stage. Drawing inspiration from the NEFTUNE method, which highlights the potential performance benefits of introducing controlled randomness during language model training. We adopt a similar principle in our approach~\cite{noise}. Additionally, echoing the compression method utilized in BERT-of-Theseus, the authors recognize that random layer replacement is an effective approach to model compression~\cite{theseus}. In our case, randomly skipping decoder layers can be seen as a form of compression as well. Our experimental results reveal that using 80\% of the decoder output mixed with 20\% of the encoder output can significantly improve performance on downstream tasks.

\section{Results}

%In order to examine the benefits of BPDec, we conduct experiments on various NLP tasks with a few models as baselines.

\subsection{Pretraining}

Pretraining was conducted on both base and large model sizes. Additionally, we introduced a variant of the base model (base-h256) to further demonstrate the generalizability of our method on smaller models. The encoder architecture, pretraining data, pretraining tasks, and mask settings are identical with the original BERT models. Our BPDec lies in the addition of a decoder following the encoder. Details about the parameters of this decoder can be found in Table~\ref{table:hyperparam}. This strategic augmentation aims to enhance the model's capabilities while maintaining the foundational architecture of BERT encoder, and it also ensures a high degree of comparativeness in our comparative experiments.

\begin{table}[ht]
\centering
\begin{tabular}{lccc}
\hline
\textbf{BERT+BPDec Hyperparam} & \textbf{base-h256} & \textbf{base} & \textbf{large} \\
\hline
Number of decoder layers & 2 & 2 & 4 \\
decoder layers with unmasking & [1, 2] & [1, 2] & [1, 3] \\
decoder layer unmasking rates & [0.5, 1.0] & [0.5, 1.0] & [0.5, 1.0] \\
\% of mix with encoder output & 20 & 20 & 20 \\
\rowcolor{gray!30} Number of encoder layers & 12 & 12 & 24 \\
\rowcolor{gray!30} Hidden size & 256 & 768 & 1024 \\
\rowcolor{gray!30} FFN inner hidden size & 1024 & 3072 & 4096 \\
\rowcolor{gray!30} Number of attention heads & 4 & 12 & 16 \\
\rowcolor{gray!30} Attention head size & 64 & 64 & 64 \\
\hline
\end{tabular}
\caption{Hyperparameters for decoders used in BERT+BPDec-base-h256, BERT+BPDec-base and BERT+BPDec-large pretraining. The architectural hyperparameters, including hidden size and the number of attention heads, are the exaclty same with the corresponding BERT model. These parameters that remain unchanged are highlighted with a gray background in the table for easy reference.}
\label{table:hyperparam}
\end{table}

In our study, we conducted a comparative analysis using the BERT model, BERT model with pre-LN (pre-layer normalization)~\cite{preLN}, and DeBERTa~\cite{deberta} trained under identical settings as baselines. Here, we utilized the same hyperparameters, tokenizer, and pretraining dataset to isolate the impact of other factors on model performance. Our focus was solely on determining the effectiveness of the architectural change. In RoBERTa~\cite{roberta}, authors significantly extended the pretraining duration, increasing the number of steps from 100K to 500K. This led to substantial improvements in downstream task performance. Authors also emphasized that the longest-trained model still haven't show signs of overfitting. This observation highlights the importance of controlling for training steps and pretraining dataset when evaluating architectural modifications, as it can act as a confounding factor that influences model performance. Without careful consideration of such factors, drawing definitive conclusions about the true impact of architectural changes could be misleading.

Moreover, it is important to note that although the DeBERTa model was trained for the same number of steps as other models, its disentangled attention mechanism introduces a significant overhead comparing with other baseline models and with our BERT+BPDec models. The comparisons can be found in Table~\ref{table:tflops} This added complexity affects both the pretraining and fine-tuning phases, but ours BERT+BPDec models only have extra ovearhead during pretraining. The implications of this increased overhead are critical, especially when considering the balance between model performance and computational efficiency. Our analysis aims to provide insights into how these different architectures impact training and computation efficiency, while also maintaining or enhancing language processing capabilities.

\begin{table}[ht]
\centering
\begin{tabular}{lccc}
\hline
\textbf{Models} & \textbf{BERT-large} & \textbf{BERT-large+BPDec} & \textbf{DeBERTa-large} \\
\hline
TFLOPs pretraining & 1.014 & 1.182 & 1.349 \\
Time costs pretraining & 100\% & +17.7\% & +37.41\% \\
TFLOPs finetuning & 1.008 & 1.176 & 1.343 \\
Time costs finetuning & 100\% & +0.0\% & +37.63\% \\
\hline
\end{tabular}
\caption{Comparison of TFLOPs and Relative Time Costs for Vanilla BERT, BERT+BPDec, and DeBERTa Models. TFLOPs are calculated per data point during pretraining and finetuning, including forward and backward propagation. Time costs are expressed as percentages relative to the Vanilla BERT model (set at 100\%).}
\label{table:tflops}
\end{table}

\subsection{Performance on GLUE Tasks}

We summarize the results on seven Natural Language Understanding (NLU) tasks from the GLUE benchmark. These tasks and their corresponding metrics are as follows:

\begin{itemize}
\item \textbf{AX (GLUE Diagnostic Task):} Tests ability to understand linguistic phenomena such as logic, predicate-argument structure, and lexical semantics.
\item \textbf{CoLA (Corpus of Linguistic Acceptability):} Measures grammatical correctness.
\item \textbf{MNLI-m/mm (Multi-Genre NLI - Matched/Mismatched):} Assesses the ability to predict textual entailment across different genres.
\item \textbf{MRPC (Microsoft Research Paraphrase Corpus):} Focuses on identifying whether two sentences are paraphrases of each other.
\item \textbf{QNLI (Question NLI):} Involves determining whether a context sentence contains the answer to a question.
\item \textbf{QQP (Quora Question Pairs):} Focuses on determining if two questions asked on Quora are semantically equivalent.
\item \textbf{RTE (Recognizing Textual Entailment):} Sentence entailment task.
\item \textbf{STS (Semantic Textual Similarity Benchmark):} Evaluates the degree of semantic similarity between two sentences.
\end{itemize}

\begin{table}[ht]
\centering
\scriptsize
\begin{tabular}{lcccccccccc}
\hline
Model & AX & COLA & MNLI-m/mm & MRPC & QNLI & QQP & RTE & SST & Avg \\
\hline
BERT-base-h256~\cite{bert} & 78.99 & 75.26 & 79.09/79.83 & \textbf{81.86} & 87.81 & 88.45 & 61.37 & 87.73 & 80.04 \\
BERT-PreLN-base-h256~\cite{preLN} & 76.70 & 69.13 & 76.48/77.87 & 71.81 & 83.40 & 86.59 & 55.96 & 86.01 & 75.99 \\
\textcolor{gray}{DeBERTa-base-h256}~\cite{deberta} & \textcolor{gray}{79.04} & \textcolor{gray}{74.50} & \textcolor{gray}{\textbf{79.41}}/\textcolor{gray}{79.99} & \textcolor{gray}{78.19} & \textcolor{gray}{87.88} & \textcolor{gray}{88.68} & \textcolor{gray}{\textbf{62.62}} & \textcolor{gray}{88.76} & \textcolor{gray}{79.89} \\
\textbf{BERT+BPDec-base-h256} & \textbf{79.06} & \textbf{75.74} & 79.39/\textbf{80.33} & \textbf{81.86} & \textbf{88.05} & \textbf{88.96} & 62.59 & \textbf{89.33} & \textbf{80.59} \\
\hline
BERT-base~\cite{bert} & 85.05 & 83.13 & 84.67/84.85 & 88.48 & 91.84 & 90.85 & 71.92 & 93.58 & 86.04 \\
BERT-PreLN-base~\cite{preLN} & 85.03 & \textbf{83.89} & 85.10/85.45 & 88.24 & 91.96 & 90.78 & 71.56 & 92.89 & 86.10 \\
\textcolor{gray}{DeBERTa-base}~\cite{deberta} & \textcolor{gray}{85.43} & \textcolor{gray}{82.74} & \textcolor{gray}{85.51}/\textcolor{gray}{\textbf{85.66}} & \textcolor{gray}{88.48} & \textcolor{gray}{92.17} & \textcolor{gray}{\textbf{91.18}} & \textcolor{gray}{\textbf{71.79}} & \textcolor{gray}{93.58} & \textcolor{gray}{86.28} \\
\textbf{BERT+BPDec-base} & \textbf{85.76} & 83.13 & \textbf{85.66}/85.59 & \textbf{88.73} & \textbf{92.39} & 90.98 & 71.12 & \textbf{93.81} & \textbf{86.35} \\
\hline
BERT-large~\cite{bert} & 87.21 & 85.43 & 86.32/87.49 & 87.50 & 93.47 & 91.44 & 72.92 & 94.04 & 87.31 \\
BERT-PreLN-large~\cite{preLN} & 87.11 & 84.99 & 86.91/87.41 & \textbf{88.24} & \textbf{93.70} & 91.24 & 75.81 & 93.69 & 87.68 \\
\textcolor{gray}{DeBERTa-large}~\cite{deberta} & \textcolor{gray}{\textbf{87.23}} & \textcolor{gray}{85.62} & \textcolor{gray}{87.01}/\textcolor{gray}{\textbf{87.54}} & \textcolor{gray}{87.75} & \textcolor{gray}{93.14} & \textcolor{gray}{91.17} & \textcolor{gray}{76.53} & \textcolor{gray}{94.82} & \textcolor{gray}{87.87} \\
\textbf{BERT+BPDec-large} & 87.22 & \textbf{86.01} & \textbf{87.76}/87.49 & \textbf{88.24} & 93.28 & \textbf{91.42} & \textbf{76.90} & \textbf{94.85} & \textbf{88.16} \\
\hline
\end{tabular}
\caption{Results on 8 GLUE benchmark tasks. Note that the results presented here are not based on publicly available checkpoints but are derived from models we trained ourselves with aligned settings. Given that DeBERTa incurs higher costs in finetuning and has more parameters, its corresponding results are highlighted in gray for comparative analysis.}
\label{table:glue1}
\end{table}

In our finetuning experiments, each model was trained on the designated training set and subsequently evaluated on the development set. Table~\ref{table:glue1} illustrates that under identical conditions and with the same number of pretraining samples, BERT with BPDec enhancement outperforms both BERT and DeBERTa models. Notably, MNLI, the largest dataset among these tasks, serves as a robust indicator of model performance. Our model competes closely with the similarly sized DeBERTa model, achieving slightly higher scores in the matched condition and comparable results in the mismatched one. However, it's crucial to consider that the DeBERTa model's encoder is approximately 17\% larger in parameter size than BERT encoder, and it incurs 32\% additional costs in finetuning and a 29\% increase in inference latency. Despite these disparities, BERT+BPDec matches or even surpasses DeBERTa in various benchmarks in base-h256, base and large level sizes. Moreover, BERT+BPDec achieves significant improvement over the original BERT with consuming the exact same finetuning and serving costs. The results tell us that this decoder only enhancement significantly boosts performance without incurring additional costs. Building on this, we extended our testing to question answering tasks, further solidifying the conclusion that our work can yield substantial performance gains.

\subsection{Performance on SQuAD Tasks}

\begin{table}[h]
\centering
\begin{tabular}{lcccc}
\hline
& \multicolumn{2}{c}{\textbf{SQuAD v1}} & \multicolumn{2}{c}{\textbf{SQuAD v2}} \\
\textbf{Model} & \textbf{EM} & \textbf{F1} & \textbf{EM} & \textbf{F1} \\
\hline
BERT-base-h256~\cite{bert} & 75.05 & 83.55 & 67.07 & 69.75 \\
BERT-PreLN-base-h256~\cite{preLN} & 74.18 & 80.26 & 65.39 & 67.96 \\
\textcolor{gray}{DeBERTa-base-h256}~\cite{deberta} & \textcolor{gray}{\textbf{75.91}} & \textcolor{gray}{\textbf{84.31}} & \textcolor{gray}{67.32} & \textcolor{gray}{69.87}  \\
BERT+BPDec-base-h256 & 75.83 & 84.24 & \textbf{68.00} & \textbf{70.12} \\
\hline
BERT-base~\cite{bert} & 83.70 & 90.63 & 76.83 & 79.78 \\
BERT-PreLN-base~\cite{preLN} & 83.65 & 90.58 & 77.03 & 79.87 \\
\textcolor{gray}{DeBERTa-base}~\cite{deberta} & \textcolor{gray}{83.75} & \textcolor{gray}{90.78} & \textcolor{gray}{77.87} & \textcolor{gray}{81.04} \\
BERT+BPDec-base & \textbf{84.30} & \textbf{91.22} & \textbf{78.21} & \textbf{81.43} \\
\hline
BERT-large~\cite{bert} & 86.58 & 92.92 & 82.06 & 85.22 \\
BERT-PreLN-large~\cite{preLN} & 85.94 & 92.56 & 81.19 & 84.42 \\
\textcolor{gray}{DeBERTa-large}~\cite{deberta} & \textcolor{gray}{\textbf{86.91}} & \textcolor{gray}{\textbf{92.96}} & \textcolor{gray}{82.09} & \textcolor{gray}{85.12} \\
BERT+BPDec-large & 86.81 & 92.90 & \textbf{82.62} & \textbf{85.59} \\
\hline
\end{tabular}%
\caption{Results on SQuAD v1 and v2 tasks. Given that DeBERTa incurs higher costs in finetuning and has more parameters, its corresponding results are highlighted in gray for comparative analysis.}
\label{table:squad}
\end{table}

The Stanford Question Answering Dataset (SQuAD) serves as a benchmark for evaluating question answering systems in NLP. It comprises two versions: SQuAD v1, focusing on answerable questions from given passages, and SQuAD v2, which includes both answerable and unanswerable questions. Performance is measured using Exact Match (EM) and F1 Score, reflecting the precision and accuracy of the model's responses.

The experiments conducted on both the v1 and v2 tasks, as presented in Table~\ref{table:squad}, following a similar trend to those performed on the GLUE tasks: all models were trained on the training set and evaluated on the development set. While the results for BERT+BPDec on SQuAD v1 did not achieve the highest scores in the table, its performance on SQuAD v2, particularly for the base and base-h256 size model, is notably stronger than the original BERT and even surpases the DeBERTa model. These results underscore BPDec's efficiency and efficacy, showcasing its ability to deliver robust performance in question answering tasks, a key area in NLP, without additional overhead.

\subsection{Ablation Study}

The BPDec involves multiple modifications of architecture and randomization which introduce many parameters in the process. It is essential to clearly articulate whether each change is necessary and how much it contributes to the final result. Furthermore, understanding the influence of each major parameter on the outcome is also crucial. To further demonstrate the effectiveness of our method, we conducted several ablation experiments on selected GLUE tasks and SQuAD tasks with base-level sized models. For these experiments, we strategically selected GLUE tasks that have larger datasets to ensure robustness and credibility in our results and to fit the data representations within a single table effectively. Tasks like RTE and SST, which have relatively smaller datasets were ommited. In contrast, tasks such as MNLI, with its substantial dataset of nearly 400,000 examples, were included. This selection criteria is designed to illustrate the impact of our modifications more clearly and designed to be fitted into a single table. By focusing on tasks with larger datasets, we can provide more significant insights, but can still validate the efficacy of our approach comprehensively.

\subsubsection{Number of MLM Decoder Layers}

\begin{table}[h]
\centering
\scriptsize
\begin{tabular}{lccccccc}
\hline
Model & \textbf{AX} & \textbf{COLA} & \textbf{MNLI-m/mm} & \textbf{MRPC} & \textbf{QNLI} & \textbf{QQP} & \textbf{SQuAD v2 EM/F1} \\
\hline
BERT-base & 85.05 & 83.13 & 84.67/84.85 & 88.48 & 91.84 & 90.85 & 76.83/79.78 \\
+ 1 decoder layers & 84.89 & 83.28 & \textbf{85.30}/85.06 & 88.24 & \textbf{91.67} & 90.78 & 76.92/80.12 \\
\textbf{+ 2 decoder layers} & \textbf{85.10} & \textbf{83.41} & 84.82/85.53 & \textbf{88.97} & 91.31 & \textbf{90.99} & \textbf{77.67}/\textbf{81.13} \\
+ 3 decoder layers & 84.91 & \textbf{83.41} & 84.48/84.53 & 87.75 & 91.62 & 90.78 & 77.14/80.19 \\
+ 4 decoder layers & 84.74 & 81.21 & 84.19/\textbf{84.64} & 88.24 & 91.01 & 91.11 & 75.89/78.72 \\
+ 6 decoder layers & 83.49 & 80.35 & 83.98/84.19 & 86.52 & 91.01 & 90.81 & 74.21/77.03 \\
\hline
\end{tabular}
\caption{Comparative results on various tasks with different numbers of decoder layers added to BERT-base.}
\label{table:bert_decoder_layers}
\end{table}

In our study, we initially sought to determine the optimal number of MLM decoder layers for improving model performance. To achieve this, we conducted an ablation study focusing solely on increasing the number of decoder layers during pretraining, without implementing other changes. During finetuning, we use the same number of layers as the BERT-base model.

Our findings in Table~\ref{table:bert_decoder_layers} revealed that adding two decoder layers to the BERT-base model yielded the best results. Parallel experiments were conducted on the larger model as well, where adding four decoder layers to the BERT-large model proved the most effective. These results led us to conclude that augmenting the model with additional decoder layers equivalent to approximately one-sixth of the number of encoder layers seems to be the optimal strategy.

\subsubsection{Decoder Layers with Gradual Unmasking Attention (GUA)}

\begin{table}[h]
\centering
\scriptsize
\begin{tabular}{lccccccc}
\hline
GUA unmasking rates & \textbf{AX} & \textbf{COLA} & \textbf{MNLI-m/mm} & \textbf{MRPC} & \textbf{QNLI} & \textbf{QQP} & \textbf{SQuAD v2 EM/F1} \\
\hline
$[0.0, 0.0]$ & \textbf{85.10} & 83.41 & 84.82/\textbf{85.53} & 88.97 & 91.31 & \textbf{90.99} & 77.67/81.13 \\
$[1.0, N/A]$ & 84.77 & 83.24 & 85.01/85.08 & 87.50 & 91.32 & 90.86 & 77.68/80.82 \\
$[0.0, 1.0]$ & 84.95 & \textbf{83.70} & 84.49/85.02 & 87.75 & 91.43 & 90.86 & 78.23/81.26 \\
\textbf{$[0.5, 1.0]$} & 85.08 & 83.13 & \textbf{85.08}/85.40 & \textbf{88.46} & \textbf{91.62} & \textbf{90.99} & \textbf{78.83}/\textbf{81.73} \\
\hline
\end{tabular}
\caption{Comparative results on various tasks for a 12-layer BERT + 2-layer decoder model with different configurations on Gradual Unmasking Attention (GUA) in each layer regarding the unmasking rates during pretraining. The configurations are denoted by lists, where each list specifies 2 random unmasking rates for the GUA in the first and the second decoder layers}
\label{table:attention_mask_config}
\end{table}

\begin{figure}[ht]
  \includegraphics[width=120mm]{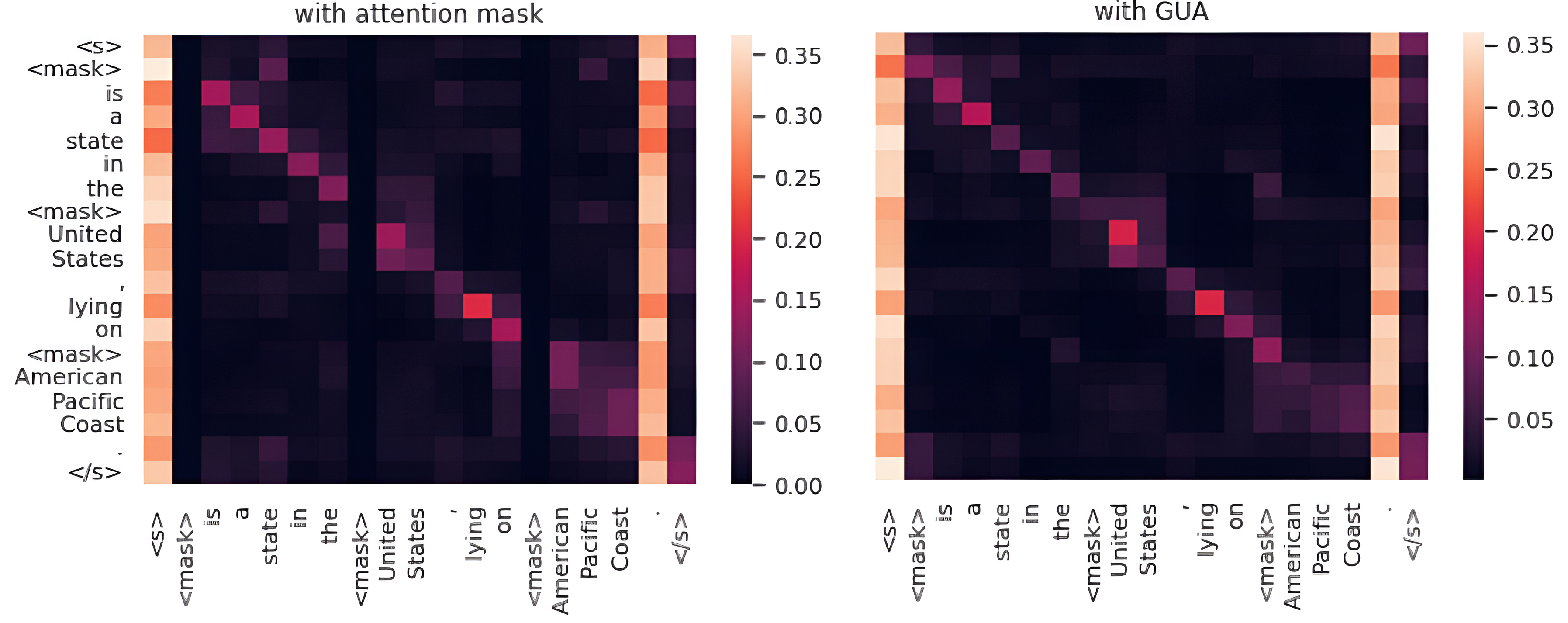}
  \centering
  \caption{Attention heatmap comparing masked attention and our GUA mechanism in the last layer of a pretrained BERT+BPDec model.}
  \label{fig:attn_heatmap}
\end{figure}

In our ablation study shown in Table~\ref{table:attention_mask_config}, we examined the effects of releasing attention mask restrictions across various configurations of the BPDec model during the MLM pretraining. Initially, the complete removal of the attention mask for the decoder under the conventional setting showed no significant impact on performance. However, significant enhancements were observed when the Gradual Unmasking Attention (GUA) strategy was applied to a deeper layer of the BPDec decoder. The optimal performance was achieved with a specific implementation of GUA: in the first decoder layer, 50\% of the positions were randomly unmasked, and in the second layer, the remaining 50\% were fully revealed. This configuration of GUA yielded the most substantial improvements across our tests. These findings underscore the intricate relationship between model architecture and the dynamic application of attention mechanisms, highlighting the transformative potential of GUA in refining MLM decoders for complex language processing tasks. The Figure~\ref{fig:attn_heatmap} is a heatmap illustrating the average attention weights over all attention heads in the last layer of a pretrained BERT+BPDec model. The visualization highlights how certain attention heads could attend to masked token embeddings, comparing the traditional masked attention mechanism with our proposed Gated Universal Attention (GUA) approach. Moreover, the enhancements on the SQuAD task, observed after integrating GUA, were significantly larger compared to those on other classification tasks. This indicates that GUA particularly boosts the model’s ability to process and understand longer and more complex textual contexts, affirming its efficacy in tasks requiring deep contextual comprehension.

\subsubsection{Effectiveness of Output Random Mix}
% \begin{table}[h]
% \centering
% \scriptsize
% \begin{tabular}{lccccccc}
% \hline
% \textbf{Mix Axis} & \textbf{AX} & \textbf{COLA} & \textbf{MNLI-m/mm} & \textbf{MRPC} & \textbf{QNLI} & \textbf{QQP} & \textbf{SQuAD v2 EM/F1} \\
% \hline
% (batch, seq, hidden) & 84.68 & 83.03 & 84.78/84.91 & 88.73 & 91.16 & 90.80 & 77.83/80.84 \\
% \textbf{(batch, seq)} & \textbf{85.86} & 83.13 & \textbf{85.66}/\textbf{85.59} & 88.73 & \textbf{92.39} & \textbf{90.98} & \textbf{77.87}/80.94 \\
% (batch) & 85.17 & \textbf{83.60} & 85.11/85.23 & \textbf{88.97} & 92.46 & 90.88 & 78.37/\textbf{81.34} \\
% \hline
% \end{tabular}
% \caption{Comparative results on various tasks with different mixing axis configurations on BERT+BPDec-base model during pretraining. We employ the decoder output values 80\% of the time, while for the remaining 20\%, we utilize the values from the encoder. The configurations specify the dimensions across which mixing is applied, impacting the model's performance on a range of tasks.}
% \label{table:mixing_axis_config}
% \end{table}

\begin{table}[h]
\centering
\scriptsize
\begin{tabular}{lccccccc}
\hline
\textbf{Ratio of Decoder} & \textbf{AX} & \textbf{COLA} & \textbf{MNLI-m/mm} & \textbf{MRPC} & \textbf{QNLI} & \textbf{QQP} & \textbf{SQuAD v2 EM/F1} \\
\hline
0.5 & 84.64 & 83.22 & 84.51/85.28 & 87.99 & 92.17 & \textbf{91.09} & 76.22/79.29 \\
0.6 & 84.65 & 82.45 & 85.04/85.05 & 87.75 & 91.93 & 90.99 & 77.43/80.45 \\
0.7 & 84.81 & \textbf{83.89} & 85.08/85.15 & 87.99 & 92.37 & 91.08 & \textbf{78.31}/\textbf{81.49} \\
\textbf{0.8} & \textbf{85.76} & 83.13 & \textbf{85.66}/\textbf{85.59} & \textbf{88.73} & \textbf{92.39} & 90.98 & 78.11/81.23 \\
1.0 & 85.08 & 83.13 & 85.08/85.40 & 88.46 & 91.62 & 90.99 & \textbf{78.83}/\textbf{81.73} \\
\hline
\end{tabular}
\caption{Comparative results on various tasks with different ratios of decoder output in a BERT+BPDec-base model during pretraining.}
\label{table:decoder_output_ratio}
\end{table}

In our ablation studies, we explored the integration of randomness into the model output before the Softmax layer with Masked Language Modeling Task. A particularly effective method we found is to randomly alternate between using outputs from the decoder and the encoder. We experimented with varying mix ratios of decoder and encoder outputs. As shown in Table~\ref{table:decoder_output_ratio}, the most effective ratio during pretraining was found to be 80\% decoder output and 20\% encoder output. This specific mix ratio led to the most balanced performance, underlining the importance of fine-tuning the balance between the two types of outputs to achieve optimal results in the finetuning phase. The 80\% ratio for utilizing decoder output emerged as a reasonable choice. If this ratio were lower, it might result in insufficient training of the decoder, diminishing its potential benefits. Conversely, if the ratio were higher, it could undermine the intended effect of randomness. 

From the perspective of experimental results, although completely avoiding random output mix yielded the best performance on the SQuAD task, random output mix provides us with an excellent method to balance the model's capabilities across different tasks. We believe that the 80\% rate trades off a portion of SQuAD's context understanding ability in exchange for better performance on other GLUE tasks, ultimately achieving a good balance. Therefore, this balanced approach not only ensures effective training of the decoder but also maintains the necessary level of randomness to optimize the model's overall performance.

\subsection{Generalizability to Other BERT-like Models}

\begin{table}[h]
\centering
\scriptsize
\begin{tabular}{lccccccc}
\hline
\textbf{Model} & \textbf{AX} & \textbf{COLA} & \textbf{MNLI-m/mm} & \textbf{MRPC} & \textbf{QNLI} & \textbf{QQP} & \textbf{SQuAD v2 EM/F1} \\
\hline
ALBERT-base & 84.34 & 80.15 & 84.46/84.51 & 85.78 & 91.03 & 90.81 & 75.62/78.77 \\
+BPDec & \textbf{84.43} & \textbf{81.78} & \textbf{84.82/85.09} & 87.75 & \textbf{92.04} & \textbf{90.90} & \textbf{76.71/79.93} \\
\hline
BERT-PreLN-base & 85.03 & 83.89 & 85.10/85.45 & 88.24 & 91.96 & 90.78 & 77.03/79.87 \\
\textbf{+BPDec} & \textbf{85.42} & \textbf{85.08} & \textbf{85.55/85.86} & \textbf{88.75} & 91.71 & 90.77 & \textbf{78.04/80.92} \\
\hline
\hline
DeBERTa-base & 85.43 & 82.74 & 85.51/85.86 & 88.48 & 92.17 & 91.18 & 78.21/81.43 \\
\textbf{+BPDec} & 85.43 & \textbf{83.31} & 85.49/\textbf{85.91} & \textbf{88.56} & 91.98 & \textbf{91.21} & \textbf{78.60/81.56} \\
\hline
\end{tabular}
\caption{Comparative results of ALBERT, BERT-PreLN and DeBERTa models with and without BPDec. For DeBERTa-base, +BPDec means adding GUA and Random Mix to the original DeBERTa decoder.}
\label{table:generalize}
\end{table}

To further demonstrate the effectiveness of BPDec, we implemented and applied the same methodology to ALBERT~\cite{albert}, BERT-PreLN~\cite{preLN} and DeBERTa~\cite{deberta} models. For ALBERT implementation, the BPDec does not share parameters with the ALBERT encoder, but the decoder layers within BPDec share parameters. For BERT-preLN, we followed the encoder's structure and used pre-LayerNorm in BPDec, ensuring the structural parity between encoder and decoder. Experiments on the GLUE benchmark and SQuAD dataset shows consistent performance improvements across the models. For DeBERTa model, because the original model already has a similar decoder structure, we added GUA and applied random mix upon the original decoder.

These additional experiments shown in~\ref{table:generalize} prove the effectiveness, adaptability, and generalizability of the BPDec method, highlighting its potential for broader application in various BERT-like architectures. The consistent improvements observed across different model types suggest that BPDec's principles can be successfully transferred to enhance the performance of a wider range of models.

\section{Conclusion}

In this paper, we introduced BERT+BPDec, an enhanced version of the original BERT model, focusing on improvements with the MLM Decoder used in pretraining. Key innovations include the additions of transformer blocks as an MLM decoder, the removal of restrictions on attending to masked positions via the novel Gradual Unmasking Attention (GUA), and the introduction of output randomness by randomly mixing encoder and decoder outputs. These enhancements allowed the BERT+BPDec model to demonstrate superior performance over both the original BERT and other state-of-the-art models in rigorous evaluations on tasks such as MNLI, SQuAD, and RACE, without increasing computational complexity during further finetuning and inference. The BPDec not only facilitates improved performance and efficiency but also contributes to sustainable computing by reducing energy consumption and emissions. Moreover, the ablation study underscored the effectiveness of each modification and the optimal hyperparameter settings are provided. This work has practical implications for real-world applications of language models and paves new avenues for future research on optimizing the language modeling. While our research currently only focuses on enhancing BERT and similar MLM-based models, we are enthusiastic about its potential to inspire adaptations in general language modeling that could benefit models like GPT and T5. Our work underscores a commitment to reducing computational waste and emissions across models and their applications, setting a precedent for environmentally conscious innovations in natural language processing.

\bibliographystyle{unsrt}
\bibliography{reference}
\end{document}